\documentclass[11pt,a4paper]{article}
\usepackage[hyperref]{acl2020}
\usepackage{times}
\usepackage{latexsym}
\usepackage{booktabs}

\usepackage{microtype}
\usepackage{amsfonts}
\usepackage{amssymb}
\usepackage{pifont}
\usepackage{graphicx}

\aclfinalcopy 

\newcommand{\xmark}{\ding{55}}

\def \oxford{$^\ddag$}
\def \deepmind{$^\diamond$}
\def \ucl{$^\dagger$}
\def \alan{$^\ast$}

\title{A Survey on Contextual Embeddings}

\author{
Qi Liu\oxford,
Matt J. Kusner\ucl\alan,
Phil Blunsom\oxford\deepmind,\\
\oxford University of Oxford \deepmind DeepMind \\
\ucl University College London \alan The Alan Turing Institute\\
\oxford\texttt{\{firstname.lastname\}@cs.ox.ac.uk} \\ \ucl\texttt{m.kusner@ucl.ac.uk}
}

\begin{document}
\maketitle

\begin{abstract}
Contextual embeddings, such as ELMo and BERT, move beyond global word representations like Word2Vec and achieve ground-breaking performance on a wide range of natural language processing tasks. Contextual embeddings assign each word a representation based on its context, thereby capturing uses of words across varied contexts and encoding knowledge that transfers across languages. In this survey, we review existing contextual embedding models, cross-lingual polyglot pre-training, the application of contextual embeddings in downstream tasks, model compression, and model analyses.
\end{abstract}

\section{Introduction}
Distributional word representations \cite{turian2010word,mikolov2013efficient,pennington2014glove} trained in an unsupervised manner on large-scale corpora are widely used in modern natural language processing systems. However, these approaches only obtain a single global representation for each word, ignoring their context. Different from traditional word representations, contextual embeddings move beyond word-level semantics in that each token is associated with a representation that is a function of the entire input
sequence. These context-dependent representations can capture many syntactic and semantic properties of words under diverse linguistic contexts. Previous work \cite{peters2018deep,devlin2018bert,yang2019xlnet,raffel2019exploring} has shown that contextual embeddings pre-trained on large-scale unlabelled corpora achieve state-of-the-art performance on a wide range of natural language processing tasks, such as text classification, question answering and text summarization. Further analyses \cite{liu2019linguistic,hewitt-liang-2019-designing,hewitt2019structural,tenney2019bert} demonstrate that contextual embeddings are capable of learning useful and transferable representations across languages. 

The rest of the survey is organized as follows. In Section \ref{sec:preliminary}, we define the concept of contextual embeddings. In Section \ref{sec:train_method}, we introduce existing methods for obtaining contextual embeddings. In Section \ref{sec:cross_lingual_method}, we present the pre-training methods of contextual embeddings on multi-lingual corpora. In Section \ref{sec:transfer_method}, we describe methods for applying pre-trained contextual embeddings in downstream tasks. In Section \ref{sec:compression_method}, we detail model compression methods. In Section \ref{sec:analysis}, we survey analyses that have aimed to identify the linguistic knowledge learned by contextual embeddings. We conclude the survey by highlighting some challenges for future research in Section \ref{sec:challenges}. 



\section{Token Embeddings \label{sec:preliminary}}
Consider a text corpus that is represented as a sequence $\mathcal{S}$ of tokens, $(t_1, t_2, ..., t_N)$. Distributed representations of words \cite{harris1954distributional,bengio2003neural} associate each token $t_i$ with a dense feature vector $\mathbf{h}_{t_i}$. Traditional \emph{word embedding} techniques aim to learn a global word embedding matrix $\mathbf{E} \in \mathbb{R}^{V \times d}$, where $V$ is the vocabulary size and $d$ is the number of dimensions. Specifically, each row $\mathbf{e}_i$ of $\mathbf{E}$ corresponds to the global embedding of word type $i$ in the vocabulary $V$. Well-known models for learning word embeddings include Word2vec \cite{mikolov2013efficient} and Glove \cite{pennington2014glove}. 
On the other hand, methods that learn \emph{contextual embeddings} associate each token $t_i$ with a representation that is a function of the entire input sequence $\mathcal{S}$, i.e.\ $\mathbf{h}_{t_i} = f(\mathbf{e}_{t_1}, \mathbf{e}_{t_2}, ..., \mathbf{e}_{t_N})$, where each input token $t_j$ is usually mapped to its non-contextualized representation $\mathbf{e}_{t_j}$ first, before applying an aggregation function $f$. These context-dependent representations are better suited to capture sequence-level semantics (e.g.\ polysemy) than non-contextual word embeddings. There are many model architectures for $f$, which we review here. We begin by describing pre-training methods for learning contextual embeddings that can be used in downstream tasks.

\section{Pre-training Methods for Contextual Embeddings\label{sec:train_method}}

In large part, pre-training contextual embeddings can be divided into either unsupervised methods (e.g.\ language modelling and its variants) or supervised methods (e.g.\ machine translation and natural language inference).

\subsection{Unsupervised Pre-training via Language Modeling}

The prototypical way to learn distributed token embeddings is via language modelling. A language model is a probability distribution over a sequence of tokens. Given a sequence of $N$ tokens, $(t_1, t_2, ..., t_N)$, a language model factorizes the probability of the sequence as:
\begin{equation}
    p(t_1, t_2, ..., t_N) = \prod_{i=1}^N p(t_i | t_1, t_2, ..., t_{i-1}).
\end{equation}
Language modelling uses maximum likelihood estimation (MLE), often penalized with regularization terms, to estimate model parameters. A left-to-right language model takes the left context, $t_1, t_2, ..., t_{i-1}$, of $t_i$ into account for estimating the conditional probability. Language models are usually trained using large-scale unlabelled corpora. The conditional probabilities are most commonly learned using neural networks \cite{bengio2003neural}, and the learned representations have been proven to be transferable to downstream natural language understanding tasks \cite{dai2015semi,ramachandran2016unsupervised}.


\paragraph{Precursor Models.}
Dai and Le \shortcite{dai2015semi} is the first work we are aware of that uses language modelling together with a sequence autoencoder to improve sequence learning with recurrent networks. Thus, it can be thought of as a precursor to modern contextual embedding methods. Pre-trained on the datasets IMDB, Rotten Tomatoes, 20 Newsgroups, and DBpedia, the model is then fine-tuned on sentiment analysis and text classification tasks, achieving strong performance compared to randomly-initialized models.

Ramachandran et al. \shortcite{ramachandran2016unsupervised} extends Dai and Le \shortcite{dai2015semi} by proposing a pre-training method to improve the accuracy of sequence to sequence (seq2seq) models. The encoder and decoder of the seq2seq model is 
initialized with the pre-trained weights of two language models. These language models are separately trained on either the News Crawl English or German corpora for machine translation, while both are initialized with the language model trained with the English Gigaword corpus for abstractive summarization. These pre-trained models are fine-tuned on the WMT English $\rightarrow$ German task and the 
CNN/Daily Mail corpus, respectively, achieving better results over baselines without pre-training.

The work in the following sections improves over Dai and Le \shortcite{dai2015semi} and Ramachandran et al.\ \shortcite{ramachandran2016unsupervised} with new architectures (e.g.\ Transformer), larger datasets, and new pre-training objectives. A summary of the models and the pre-training objectives is shown in Table \ref{tab:model} and \ref{tab:objective}.

\begin{table*}[h]
    \centering
    \small
    \begin{tabular}{lcccccc}
        \toprule
            \textbf{Method} & \textbf{Architecture} & \textbf{Encoder} & \textbf{Decoder} & \textbf{Objective} & \textbf{Dataset}\\ \midrule
            ELMo & LSTM & \xmark & \checkmark & LM & 1B Word Benchmark\\
            GPT & Transformer & \xmark & \checkmark & LM & BookCorpus \\
            GPT2 & Transformer & \xmark & \checkmark & LM & Web pages starting from Reddit\\
            BERT & Transformer & \checkmark & \xmark & MLM \& NSP & BookCorpus \& Wiki\\
            RoBERTa & Transformer & \checkmark& \xmark & MLM & \scriptsize{BookCorpus, Wiki, CC-News, OpenWebText, Stories}\\
            ALBERT & Transformer & \checkmark & \xmark & MLM \ \& SOP & Same as RoBERTa and XLNet\\
            UniLM & Transformer & \checkmark & \xmark & LM, MLM, seq2seq LM & Same as BERT\\
            ELECTRA & Transformer & \checkmark & \xmark & Discriminator (o/r) & Same as XLNet\\
            XLNet & Transformer & \xmark & \checkmark & PLM & \scriptsize{BookCorpus, Wiki, Giga5, ClueWeb, Common Crawl}\\
            XLM & Transformer & \checkmark & \checkmark & CLM, MLM, TLM & Wiki, parellel corpora (e.g.\ MultiUN) \\
            MASS & Transformer & \checkmark & \checkmark & Span Mask & WMT News Crawl\\
            T5 & Transformer & \checkmark & \checkmark & Text Infilling & Colossal Clean Crawled Corpus\\
            BART & Transformer & \checkmark & \checkmark & Text Infilling \& Sent Shuffling & Same as RoBERTa \\
        \bottomrule
    \end{tabular}
    \caption{A comparison of popular pre-trained models.\label{tab:model}}
\end{table*}

\begin{table*}[h]
    \centering
    \small
    \begin{tabular}{lccc}
        \toprule
            \textbf{Objective} & \textbf{Inputs} & \textbf{Targets}\\ \midrule
            LM & [START] & I am happy to join with you today\\
            MLM & I am [MASK] to join with you [MASK] & happy today\\
            NSP & Sent1 [SEP] Next Sent or Sent1 [SEP] Random Sent & Next Sent/Random Sent\\
            SOP & Sent1 [SEP] Sent2 or Sent2 [SEP] Sent1 & in order/reversed\\
            Discriminator (o/r) & I am thrilled to study with you today & o o r o r o o o\\
            PLM & happy join with & today am I to you \\
            seq2seq LM & I am happy to & join with you today \\
            Span Mask & I am [MASK] [MASK] [MASK] with you today & happy to join \\
            Text Infilling & I am [MASK] with you today & happy to join\\
            Sent Shuffling & today you am I join with happy to & I am happy to join with you today\\
            TLM & How [MASK] you [SEP] [MASK] vas-tu & are Comment\\
        \bottomrule
    \end{tabular}
    \caption{Pre-training objectives and their input-output formats. \label{tab:objective}}
\end{table*}

\paragraph{ELMo.}
The ELMo model \cite{peters2018deep} generalizes traditional word embeddings by extracting context-dependent representations from a bidirectional language model. A forward $L$-layer LSTM and a backward $L$-layer LSTM are applied to encode the left and right contexts, respectively. At each layer $j$, the contextualized representations are the concatenation of the left-to-right and right-to-left representations, obtaining $N$ hidden representations, $(\mathbf{h}_{1, j}, \mathbf{h}_{2, j}, ..., \mathbf{h}_{N, j})$, for a sequence of length $N$.

To use ELMo in downstream tasks, the $(L + 1)$-layer representations (including the global word embedding) for each token $k$ are aggregated as:
\begin{equation}
    \label{eq:elmo_feature}
    \textsc{ELMo}^{task}_k = \gamma^{task} \sum_{j=0}^L s^{task}_j \mathbf{h}_{k, j}, 
\end{equation}
where $\mathbf{s}^{task}$ are layer-wise weights normalized by the softmax used to linearly combine the $(L + 1)$-layer representations of the token $k$ and $\gamma^{task}$ is a task-specific constant.

Given a pre-trained ELMo, it is straightforward to incorporate it into a task-specific architecture for improving the performance. As most supervised models use global word representations $\mathbf{x}_k$ in their lowest layers, these representations can be concatenated with their corresponding context-dependent representations $\textsc{ELMo}^{task}_k$, obtaining $[\mathbf{x}_k; \textsc{ELMo}^{task}_k]$, before feeding them to higher layers. 

The effectiveness of ELMo is evaluated on six NLP problems, including question answering, textual entailment and sentiment analysis.

\paragraph{GPT, GPT2, and Grover.}
GPT \cite{radford2018gpt} adopts a two-stage learning paradigm: (a) unsupervised pre-training using a language modelling objective and (b) supervised fine-tuning. The goal is to learn universal representations transferable to a wide range of downstream tasks. To this end, GPT uses the BookCorpus dataset \cite{zhu2015aligning}, which contains more than 7,000 books from various genres, for training the language model. The Transformer architecture \cite{vaswani2017attention} is used to implement the language model, which has been shown to better capture global dependencies from the inputs compared to its alternatives, e.g.\ recurrent networks, and perform strongly on a range of sequence learning tasks, such as machine translation \cite{vaswani2017attention} and document generation \cite{liu2018generating}. To use GPT on inputs with multiple sequences during fine-tuning, GPT applies task-specific input adaptations motivated by  traversal-style approaches \cite{rocktaschel2015reasoning}. These approaches pre-process each text input as a single contiguous sequence of tokens through special tokens including [START] (the start of a sequence), [DELIM] (delimiting two sequences from the text input) and [EXTRACT] (the end of a sequence). GPT outperforms task-specific architectures in 9 out of 12 tasks studied with a pre-trained Transformer. 


GPT2 \cite{radford2019language} mainly follows the architecture of GPT and trains a language model on a dataset as large and diverse as possible to learn from varied domains and contexts. To do so, 
\citet{radford2019language} create a new dataset of millions of web pages named WebText, by scraping outbound links from Reddit. 
The authors argue that a language model trained on large-scale unlabelled corpora begins to learn some common supervised NLP tasks, such as question answering, machine translation and summarization, without any explicit supervision signal. To validate this, GPT2 is tested on ten datasets (e.g.\ Children's Book Test \cite{hill2015goldilocks}, LAMBADA \cite{paperno2016lambada} and CoQA \cite{reddy2019coqa}) in a zero-shot setting. GPT2 performs strongly on some tasks. For instance, when conditioned on a document and questions, GPT2 reaches an F1-score of 55 on the CoQA dataset without using any labelled training data. This matches or outperforms the performance of 3 out of 4 baseline systems. As GPT2 divides texts into bytes and uses BPE \cite{sennrich-etal-2016-neural} to build up its vocabulary (instead of using characters or words, as in previous work), it is unclear if the improved performance comes from the model or the new input representation.

Grover \cite{zellers2019defending} creates a news dataset, RealNews, from Common Crawl and pre-trains a language model for generating realistic-looking fake news that is conditioned on meta-data including domains, dates, authors and headlines. They further study discriminators that can be used to detect fake news. The best defense against Grover turns out to be Grover itself, which sheds light on the importance of releasing trained models for detecting fake news.

\paragraph{BERT.}
ELMo \cite{peters2018deep} concatenates representations from the forward and backward LSTMs without considering the interactions between the left and right contexts. GPT \cite{radford2018gpt} and GPT2 \cite{radford2019language} use a left-to-right decoder, where every token can only attend to its left context. These architectures are sub-optimal for sentence-level tasks, e.g.\ named entity recognition and sentiment analysis, as it is crucial to incorporate contexts from both directions. 

BERT proposes a masked language modelling (MLM) objective, where some of the tokens of a input sequence are randomly masked, and the objective is to predict these masked positions taking the corrupted sequence as input. BERT applies a Transformer encoder to attend to bi-directional contexts during pre-training. In addition, BERT uses a next-sentence-prediction (NSP) objective. Given two input sentences, NSP predicts whether the second sentence is the actual next sentence of the first sentence. The NSP objective aims to improve the tasks, such as question answering and natural language inference, which require reasoning over sentence pairs. 

Similar to GPT, BERT uses special tokens to obtain a single contiguous sequence for each input sequence. Specifically, the first token is always a special classification token [CLS], and sentence pairs are separated using a special token [SEP]. BERT adopts a pre-training followed by fine-tuning scheme. The final hidden state of [CLS] is used for sentence-level tasks and the final hidden state of each token is used for token-level tasks. BERT obtains new state-of-the-art results on eleven natural language processing
tasks, e.g.\ improving the GLUE \cite{wang2018glue} score to 80.5\%. 

Similar to GPT2, it is unclear exactly why BERT improves over prior work as it uses different objectives, datasets (Wikipedia and BookCorpus) and architectures compared to previous methods. For partial insight on this, we refer the readers to \cite{raffel2019exploring} for a controlled comparison between unidirectional and bidirectional models, traditional language modelling and masked language modelling using the same datasets.

\paragraph{BERT variants.}
Recent work further studies and improves the objective and architecture of BERT. 

Instead of randomly masking tokens, ERNIE \cite{sun2019ernie} incorporates knowledge masking strategies, including entity-level masking and phrase-level masking. ERNIE 2.0 \cite{sun2019ernie2} further incorporates more pre-training tasks, such as semantic
closeness and discourse relations. SpanBERT \cite{joshi2019spanbert} generalizes ERNIE to mask random spans, without referring to external knowledge. StructBERT \cite{wang2019structbert} proposes a word structural objective that randomly permutes the order of 3-grams for reconstruction and a sentence structural objective that predicts the order of two consecutive segments.

RoBERTa \cite{liu2019roberta} makes a few changes to the released BERT model and achieves substantial improvements. The changes include: (1) Training the model longer with larger batches and more data; (2) Removing the NSP objective; (3) Training on longer sequences; (4) Dynamically changing the masked positions during pre-training. 

ALBERT \cite{lan2019albert} proposes two parameter-reduction techniques (factorized embedding parameterization and cross-layer parameter sharing) to lower memory consumption and speed up training. Furthermore, ALBERT argues that the NSP objective lacks difficulty, as the negative examples are created by pairing segments from different documents, this mixes topic prediction and coherence prediction into a single task. ALBERT instead uses a sentence-order prediction (SOP) objective. SOP obtains positive examples by taking out two consecutive segments and negative examples by reversing the order of two consecutive segments from the same document.

\paragraph{XLNet.}
The XLNet model \cite{yang2019xlnet} identifies two weaknesses of BERT:

\begin{enumerate}
    \item BERT assumes conditional independence of corrupted tokens. For instance, to model the probability $p(t_2 \!=\! \textrm{cat}, t_6 \!=\! \textrm{mat} | t_1 \!=\! \textrm{The}, t_2 \!=\! \textrm{[MASK]}, t_3 \!=\! \textrm{sat}, t_4 \!=\! \textrm{on}, t_5 \!=\! \textrm{the}, t_6 \!=\! \textrm{[MASK]})$, BERT factorizes it as $p(t_2 \!=\! \textrm{cat} | ...) p(t_6 \!=\! \textrm{mat} | ...)$, where $t_2$ and $t_6$ are assumed to be conditionally independent.
    \item The symbols such as [MASK] are introduced by BERT during pre-training, yet they never occur in real data, resulting in a discrepancy between pre-training and fine-tuning.
\end{enumerate}

XLNet proposes a new auto-regressive method based on permutation language modelling (PLM) \cite{uria2016neural} without introducing any new symbols. The MLE objective for it is calculated as:
\begin{equation}
    \max_{\theta} \: \mathbb{E}_{\mathbf{z} \in Z_N} \left[ \sum_{j=1}^N \log p_{\theta}(t_{z_j} | t_{z_1}, t_{z_2}, ..., t_{z_{j-1}}) \right].
\end{equation}
For each sequence, XLNet samples a permutation order $\mathbf{z} = [z_1, z_2, ..., z_N]$ from the set of all permutations $Z_N$, where $|Z_N| = N!$. The probability of the sequence is factorized according to $\mathbf{z}$, where the $z_j$-th token $t_{z_j}$ is conditioned on all the previous tokens $t_{z_1}, t_{z_2}, ..., t_{z_j}$ according to the permutation order $\mathbf{z}$.

XLNet further adopts two-stream self-attention and Transformer-XL \cite{dai2019transformer} to take into account the target positions $z_j$ and learn long-range dependencies, respectively.

As the cardinality of $Z_N$ is factorial, naive optimization would be challenging. Thus, XLNet conditions on part of the input and generates the rest of the input  to reduce the scale of the search space:
\begin{equation}
    \max_{\theta} \: \mathbb{E}_{\mathbf{z} \in Z_N} \left[ \sum_{j=c + 1}^N \log p_{\theta}(t_{z_j} | t_{z_1}, t_{z_2}, ..., t_{z_{j-1}}) \right],
\end{equation}
where $c$ is the cutting point of the sequence. However, it is tricky to compare XLNet directly with BERT due to the multiple changes in loss and architecture.\footnote{We note that RoBERTa, which makes much smaller changes to BERT is able to outperform XLNet. Future study needs to be done to understand the precise advantages of XLNet's modifications to BERT.}



\paragraph{UniLM.}
UniLM \cite{dong2019unified} adopts three objectives: (a) language modelling, (b) masked language modelling, and (c) sequence-to-sequence language modelling (seq2seq LM), for pre-training a Transformer network. To implement three objectives in a single network, UniLM utilizes specific self-attention masks to control what context the prediction conditions on. For example, MLM can attend to its bidirectional contexts, while seq2seq LM can attend to bidirectional contexts for source sequences and left contexts only for target sequences. 

\paragraph{ELECTRA.}
Compared to BERT, ELECTRA \cite{anonymous2020electra} proposes a more effective pre-training method. Instead of corrupting some positions of inputs with [MASK], ELECTRA replaces some tokens of the inputs with their plausible alternatives sampled from a small generator network. 
ELECTRA trains a discriminator to predict whether each token in the corrupted input was replaced by the generator or not. The pre-trained discriminator can then be used in downstream tasks for fine-tuning, improving upon the pre-trained representation learned by the generator.

\paragraph{MASS.}
Although BERT achieves state-of-the-art performance for many natural language understanding tasks, BERT cannot be easily used for natural language generation. MASS \cite{song2019mass} uses masked sequences to pre-train sequence-to-sequence models. More specifically, MASS adopts an encoder-decoder framework and extends the MLM objective. The encoder takes as input a sequence where consecutive tokens are masked and the decoder predicts these masked consecutive tokens autoregressively. MASS achieves significant improvements over baselines without pre-training or with other pre-training methods on a variety of zero/low-resource language generation tasks, including neural machine translation, text summarization and conversational response generation.

\paragraph{T5.}
Raffel et al.\ \shortcite{raffel2019exploring} propose T5 (Text-to-Text Transfer Transformer), unifying natural language understanding and generation by converting the data into a text-to-text format and applying a encoder-decoder framework. 

T5 introduces a new pre-training dataset, Colossal Clean Crawled Corpus by cleaning the web pages from Common Crawl. T5 also systematically compares previous methods in terms of pre-training objectives, architectures, pre-training datasets, and transfer approaches. T5 adopts a text infilling objective (where spans of text are replaced with a single mask token), longer training, multi-task pre-training on GLUE or SuperGLUE, fine-tuning on each individual GLUE and SuperGLUE tasks, and beam search. ERNIE-GEN \cite{xiao2020ernie} is another work using text infilling, where tokens of each masked span are generated non-autoregressively.

For fine-tuning, to convert the input data into a text-to-text framework, T5 utilizes the token vocabulary of the decoder as the prediction labels. For example, the tokens ``entailment", ``contradiction", and ``neutral" are used as the labels for natural language inference tasks. For the regression task (e.g.\ STS-B \cite{cer2017semeval}), T5 simply rounds up the
scores to the nearest multiple of 0.2 and converts the results to literal string representations (e.g.\ 2.57 is converted to the string ``2.6"). T5 also adds a task-specific prefix to each input sequence to specify its task. For instance, T5 adds the prefix ``translate English to German" to each input sequence like ``That is good." for English-to-German translation datasets.

\paragraph{BART.}
The BART model \cite{lewis2019bart} introduces additional noising functions beyond MLM for pre-training sequence-to-sequence models. First, the input sequence is corrupted using an arbitrary noising function. Then, the corrupted input is reconstructed by a Transformer network trained using teacher forcing \cite{williams1989learning}. BART evaluates a wide variety of noising functions, including token masking, token deletion, text infilling, document rotation, and sentence shuffling (randomly shuffling the word order of a sentence). The best performance is achieved by using both sentence shuffling and text infilling. BART matches the performance of RoBERTa on GLUE and SQuAD and achieves state-of-the-art performance on a variety of text generation tasks.

\subsection{Supervised Objectives}

Pre-training on the ImageNet dataset (which has supervision about the objects in images) before fine-tuning on downstream tasks has become the \textit{de facto} standard in the computer vision community. Motivated by the success of supervised pre-training in computer vision, some work \cite{conneau2017supervised,mccann2017learned,subramanian2018learning} utilizes data-rich tasks in NLP to learn transferable representations.

CoVe \cite{mccann2017learned} shows that the representations learned from machine translation are transferable to downstream tasks. CoVe uses a deep LSTM encoder from
a sequence-to-sequence model trained for machine translation to obtain contextual embeddings. Empirical results show that augmenting non-contextualized word representations \cite{mikolov2013efficient,pennington2014glove} with CoVe embeddings improves performance over a wide variety of common NLP tasks, such as sentiment analysis, question classification, entailment, and question answering. InferSent \cite{conneau2017supervised} obtains contextualized representations from a pre-trained natural language inference model on SNLI.  Subramanian et al. \shortcite{subramanian2018learning} use multi-task learning to pre-train a sequence-to-sequence model for obtaining general representations, where the tasks include skip-thought \cite{kiros2015skip}, machine translation, constituency parsing, and natural language inference.

\section{Cross-lingual Polyglot Pre-training for Contextual Embeddings\label{sec:cross_lingual_method}}

Cross-lingual polyglot pre-training aims to learn joint multi-lingual representations, enabling knowledge transfer from data-rich languages like English to data-scarce languages like Romanian. Based on whether joint training and a shared vocabulary are used, we divide previous work into three categories.

\paragraph{Joint training \& shared vocabulary.} Artetxe and Schwenk \shortcite{artetxe2019massively} use a BiLSTM encoder-decoder framework with a shared BPE vocabulary for 93 languages. The framework is pre-trained using parallel corpora, including as Europarl and Tanzil. The contextual embeddings from the encoder are used to train classifiers using English corpora for downstream tasks. As the embedding space and the encoder are shared, the resultant classifiers can be transferred to any of the 93 languages without further modification. Experiments show that these classifiers achieve competitive performance on cross-lingual natural language inference, cross-lingual document classification, and parallel corpus mining. 

Rosita \cite{mulcaire2019polyglot} pre-trains a language model using text from different languages, showing the benefits of polyglot learning on low-resource languages. 

Recently, the authors of BERT developed a multi-lingual BERT\footnote{https://github.com/google-research/bert/blob/master/multilingual.md} which is pre-trained using the Wikipedia dump with more than 100 languages.

XLM \cite{lample2019cross} uses three pre-training methods for learning cross-lingual language models: (1) Causal language modelling, where the model is trained to predict $p(t_i | t_1, t_2, ..., t_{i-1})$, (2) Masked language modelling, and (3) Translation language modelling (TLM). Parallel corpora are used, and tokens in both source and target sequences are masked for learning cross-lingual association. XLM performs strongly on cross-lingual classification, unsupervised machine translation, and supervised machine translation. XLM-R \cite{conneau2019unsupervised} scales up XLM by training a Transformer-based masked language model on one hundred languages, using more than two terabytes of filtered CommonCrawl data. XLM-R shows that large-scale multi-lingual pre-training leads to significant performance gains for a wide range of cross-lingual transfer tasks.

\paragraph{Joint training \& separate vocabularies.}
Wu et al. \shortcite{wu2019emerging} study the emergence of cross-lingual structures in pre-trained multi-lingual language models. It is found that cross-lingual transfer is possible even when there is no shared vocabulary across the monolingual corpora, and there are universal latent symmetries in the embedding spaces of different languages. 

\paragraph{Separate training \& separate vocabularies.}
Artetxe et al. \shortcite{artetxe2019cross} use a four-step method for obtaining multi-lingual embeddings. Suppose we have the monolingual sequences of two languages $L_1$ and $L_2$: (1) Pre-training BERT with the vocabulary of $L_1$ using $L_1$'s monolingual data. (2) Replacing the vocabulary of $L_1$ with the vocabulary of $L_2$ and training new vocabulary embeddings, while freezing the other parameters, using $L_2$'s monolingual data. (3) Fine-tuning the BERT model for a downstream task using labeled data in $L_1$, while freezing
$L_1$'s vocabulary embeddings. (4) Replacing the fine-tuned BERT with $L_2$'s vocabulary embeddings for zero-shot transfer tasks.


\section{Downstream Learning}
\label{sec:transfer_method}

Once learned, contextual embeddings have demonstrated impressive performance when used downstream on various learning problems. Here we describe the ways in which contextual embeddings are used downstream, the ways in which one can avoid forgetting information in the embeddings during downstream learning, and how they can be specialized to multiple learning tasks.

\subsection{Ways to Use Contextual Embeddings Downstream}

There are three main ways to use pre-trained contextual embeddings in downstream tasks: (1) Feature-based methods, (2) Fine-tuning methods, and (3) Adapter methods.

\paragraph{Feature-based.} 
One example of a feature-based is the method used by ELMo \cite{peters2018deep}. Specifically, as shown in equation \ref{eq:elmo_feature}, ELMo freezes the weights of the pre-trained contextual embedding model and forms a linear combination of its internal representations. The linearly-combined representations are then used as features for task-specific architectures. The benefit of feature-based models is that they can use state-of-the-art handcrafted architectures for specific tasks.

\paragraph{Fine-tuning.} 
Fine-tuning works as follows: starting with the weights of the pre-trained contextual embedding model, fine-tuning makes small adjustments to them in order to specialize them to a specific downstream task. One stream of work applies minimal changes to pre-trained models to take full advantage of their parameters. The most straightforward way is adding linear layers on top of the pre-trained models \cite{devlin2018bert,lan2019albert}. Another method \cite{radford2019language,raffel2019exploring} uses universal data formats without introducing new parameters for downstream tasks. 

To apply pre-trained models to structurally different tasks, where task-specific architectures are used, as much of the model is initialized with pre-trained weights as possible. 
For instance, XLM \cite{lample2019cross} applies two pre-trained monolingual language models to initialize the encoder and the decoder for machine translation, respectively, leaving only cross-attention weights randomly initialized.

\paragraph{Adapters.} 
Adapters \cite{rebuffi2017learning,stickland2019bert} are small modules added between layers of pre-trained models to be trained in a multi-task learning setting. The parameters of the pre-trained model are fixed while tuning these adapter modules. Compared to previous work that fine-tunes a separate pre-trained model for each task, a model with shared adapters for all tasks often requires fewer parameters.

\subsection{Countering Catastrophic Forgetting}

Learning on downstream tasks is prone to overwrite the information from pre-trained models, which is widely known as the catastrophic forgetting \cite{mccloskey1989catastrophic,d2019episodic}. Previous work combats this by (1) Freezing layers, (2) Using adaptive learning rates, and (3) Regularization.

\paragraph{Freezing layers.} 
Motivated by layer-wise training of neural networks \cite{hinton2006fast}, training certain layers while freezing others can potentially reduce forgetting during fine-tuning. Different layer-wise tuning schedules have been studied. Long et al. \shortcite{long2015learning} freeze all layers except the top layer. Felbo et al. \shortcite{felbo2017using} use ``chain-thaw", which sequentially unfreezes and fine-tunes a layer at a time. Howard and Ruder \shortcite{howard2018universal} gradually unfreeze all layers one by one from top to bottom. Chronopoulou et al. \shortcite{chronopoulou2019embarrassingly} apply a three-stage fine-tuning schedule: (a) randomly-initialized parameters are updated for $n$ epochs, (b) the pre-trained parameters (except word embeddings) are then fine-tuned, (c) at last, all parameters are fine-tuned.

\paragraph{Adaptive learning rates.}
Another method to mitigate catastrophic forgetting is by using adaptive learning rates. As it is believed that the lower layers of pre-trained models tend to capture general language knowledge \cite{tenney2019bert}, Howard and Ruder \shortcite{howard2018universal} use lower learning rates for lower layers when fine-tuning.

\paragraph{Regularization.}
Regularization limits the fine-tuned parameters to be close to the pre-trained parameters. Wiese et al. \shortcite{wiese2017neural} minimize the Euclidean distance between the fine-tuned parameters and pre-trained parameters. Kirkpatrick et al. \shortcite{kirkpatrick2017overcoming} use the Fisher information matrix to protect the weights that are identified as essential for pre-trained models.

\subsection{Multi-task Fine-tuning}
Multi-task learning on downstream tasks \cite{liu2019multi,wang-etal-2019-tell,jozefowicz2016exploring} obtains general representations across tasks and achieves strong performance on each individual task. 

MT-DNN \cite{liu2019multi} fine-tunes BERT on all the GLUE tasks, improving the GLUE benchmark to 82.7\%. MT-DNN also demonstrates that the representations from multi-task learning obtain better performance on domain adaptation compared to BERT.

Wang et al. \shortcite{wang-etal-2019-tell} investigate further, non-GLUE tasks, such as skip-thought and Reddit response generation, for multi-task learning. 

T5 \cite{raffel2019exploring} studies various settings of multi-task learning and finds that using multi-task learning before fine-tuning on each task performs the best.
 
\section{Model Compression} 
\label{sec:compression_method}
As many pre-trained language models have a prohibitive memory footprint and latency, it is a challenging task to deploy them in resource-constrained environments. To address this, model compression \cite{cheng2017survey}, which has gained popularity in recent years for shrinking large neural networks, has been investigated for compressing contextual embedding models. Work on compressing language models utilizes (1) Low-rank approximation, (2) Knowledge distillation, and (3) Weight quantization, to make them usable in embedded systems and edge devices.

\paragraph{Low rank approximation.} 
Methods that learn low rank approximations seek to compress the full-rank model weight matrices into low-rank matrices, thereby reducing the effective number of model parameters. As the embedding matrices usually account for a large portion of model parameters (e.g.\ 21\% for $\mathrm{BERT_{Base}}$), ALBERT \cite{lan2019albert} approximates the embedding matrix $\mathbf{E} \in \mathbb{R}^{V \times d}$ as the product of two smaller matrices, $\mathbf{E}_1 \in \mathbb{R}^{V \times d'}$ and $\mathbf{E}_2 \in \mathbb{R}^{d' \times d}$, where $d' \ll d$. 

\paragraph{Knowledge distillation.}
A method called `knowledge distillation' was proposed by Hinton et al. \shortcite{hinton2015distilling}, where the `knowledge' encoded in a teacher network is transferred to a student network. Hinton et al. \shortcite{hinton2015distilling} use the soft target probabilities, output by the teacher network, to train the student network using the cross-entropy loss. The student network is smaller than the teacher network, resulting in a more lightweight model that nears the accuracy of the heavyweight teacher network. Tang et al. \shortcite{tang2019distilling} distill the knowledge from BERT into a single-layer BiLSTM, obtaining performance comparable to ELMo with roughly 100 times fewer parameters. DistilBERT \cite{sanh2019distilbert} uses MLM, distillation loss \cite{hinton2015distilling}, and cosine similarity between the embedding matrices of the teacher and student networks to train a smaller BERT model. BERT-PKD \cite{sun2019patient} uses a student BERT model with fewer layers compared to $\mathrm{BERT_{Base}}$ or $\mathrm{BERT_{Large}}$ and proposes two ways (learning from the last $k$ layers and learning from every $k$ layers) to map the layers of the student to the layers of $\mathrm{BERT_{Base}}$ or $\mathrm{BERT_{Large}}$. The hidden states of the student are kept close to the hidden states of the teacher from corresponding layers using a Euclidean distance regularizer. TinyBERT \cite{jiao2019tinybert} introduces a two-stage learning framework, where distillation is performed at both the pre-training and the fine-tuning stages. 

\paragraph{Weight quantization.}
Quantization methods focus on mapping weight parameters to low-precision integers and floating-point numbers. Q-BERT \cite{shen2019q} proposes a group-wise quantization scheme, where the parameters are divided into groups based on attention heads, and uses a Hessian-based, mixed-precision method to compress the model.

\section{Analyzing Contextual Embeddings}
\label{sec:analysis}

While contextual embedding methods have impressive performance on a variety of natural language tasks, it is often unclear exactly why they work so well. To study this, work so far has used (1) Probe classifiers, and (2) Visualization.

\paragraph{Probe classifiers.} 
A large body of work studies contextual embeddings using \emph{probes}. These are constrained classifiers designed to explore whether syntactic and semantic information is encoded in these representations or not. 

Liu et al. \shortcite{liu2019linguistic} design a series of token labelling, segmentation, and pairwise relation tasks for studying the effectiveness of contextual embeddings. Contextual embeddings achieve competitive results compared to the state-of-the-art models on most tasks, yet fail on some fine-grained linguistic tasks (e.g.\ conjunct identification).

Hewitt and Manning \shortcite{hewitt2019structural} propose a structural probe for finding syntax in contextual embeddings. The model attempts to learn a linear transformation under which the L2 distances between tokens encode the
distances between these tokens in syntactic parsing trees like dependency trees.

Tenney et al. \shortcite{tenney2019bert} find that BERT rediscovers the traditional NLP pipeline in an interpretable and
localizable way. Specifically, it is capable at POS tagging, parsing, NER, semantic
roles, and coreference, and these are learned in order.

Jawahar et al. \shortcite{jawahar2019does} use ten sentence-level probing tasks (e.g.\ SentLen, TreeDepth) and find that BERT captures phrase-level information in earlier layers and long-distance dependency information in deeper layers.

\paragraph{Visualization.}
Another body of work uses visualization to analyze attention and fine-tuning procedures, among others.

Hao et al. \shortcite{hao2019visualizing} visualize loss landscapes and optimization trajectories when fine-tuning BERT. The visualizations show that BERT reaches a good initial point during pre-training for downstream
tasks, which can lead to better optima compared to randomly-initialized models.

Kovaleva et al. \shortcite{kovaleva2019revealing} visualize the attention heads of BERT, discovering a limited set of attention patterns across different heads. This leads to the fact that the heads of BERT are highly redundant. After manually disabling certain attention heads, better performance is obtained compared to the fine-tuned BERT models that use the full set of attention heads.

Coenen et al. \shortcite{coenen2019visualizing} visualize and analyze the geometry of BERT embeddings, finding that BERT distinguishes word senses at a very fine-grained level. These word senses are also found to be encoded in a relatively low-dimensional subspace.

\section{Current Challenges}
\label{sec:challenges}

There are many key challenges that, if solved, would improve future contextual embeddings.

\paragraph{Better pre-training objectives.} 
BERT designed MLM to take advantage of bi-directional information during pre-training. It remains unclear whether there are pre-training objectives that are simultaneously more efficient and effective. Some recent work focuses on designing new training methods \cite{anonymous2020electra}, noise combination techniques, \cite{lewis2019bart} and multi-task learning approaches \cite{wang-etal-2019-tell}. 

\paragraph{Understanding the knowledge encoded in pre-trained models.}
As described above, a range of methods \cite{tenney2019bert,tenney2019you,hewitt2019structural,liu2019linguistic} have been proposed to explore the effectiveness of pre-trained models via probes. Yet, controlled experiments are still lacking to understand whether the representations actually encode linguistic knowledge or the probes happen to learn to perform well on these linguistic tasks because the data they use is so high-dimensional \cite{hewitt-liang-2019-designing}. Hewitt and Liang \shortcite{hewitt-liang-2019-designing} devise control tasks, where a good probe is one that performs well on linguistic tasks, and badly on control tasks. They find that most existing probes fail to satisfy this condition. Indeed, most probes use shallow classifiers, which may not be able to extract the relevant information from contextual representations. New probes or better methods for understanding contextual representations are needed.



\paragraph{Model robustness.}
Concerns about the vulnerability of models to attack  are growing when deploying NLP models into production. Wallace et al. \shortcite{wallace2019universal} show that universal adversarial triggers that cause significant performance deterioration of pre-trained models can be found. Additionally, concerns of abusing pre-trained models (e.g.\ generating fake news) have arisen\footnote{https://www.theverge.com/2019/11/7/20953040/openai-text-generation-ai-gpt-2-full-model-release-1-5b-parameters}. Better methods for increasing model robustness are highly needed. 

\paragraph{Controlled generation of sequences.}
Pre-trained language models \cite{radford2018gpt,radford2019language} are able to generate realistic-looking text sequences. Yet it is hard to adapt these models to generate domain-specific sequences \cite{keskar2019ctrl} or to agree with common human knowledge \cite{zellers2019defending}. As a result, we advocate research on more fine-grained control over sequence generation. 

\section*{Acknowledgements}
We thank Douwe Kiela, Jiatao Gu, Yi Tay, Xiaodong Liu, Ziyang Wang and Jake Zhao for their comments and discussions on this manuscript. 

\bibliography{references}
\bibliographystyle{acl_natbib}

\end{document}